\documentclass{article}
\usepackage{spconf}

\usepackage{enumitem}
\setlist{nosep, leftmargin=14pt}

\usepackage{mwe} 
\usepackage{cite}
\usepackage{bm}
\usepackage{booktabs}
\usepackage{amsmath,amssymb,amsfonts}
\usepackage{algorithm}
\usepackage{algorithmic}
\usepackage{graphicx}
\usepackage{multirow}
\usepackage[normalem]{ulem}
\usepackage{textcomp}
\usepackage{xcolor}
\usepackage{colortbl}
\usepackage{lineno}

\newcommand{\figref}[1]{Fig.~\ref{#1}}

\newcommand{\tabref}[1]{Table~\ref{#1}}

\newcommand{\equaref}[1]{Eq.~\eqref{#1}}
\newcommand{\lrparen}[1]{\left( #1 \right)}

\newcommand{\lrbracket}[1]{\left< #1 \right>}

\def\BibTeX{{\rm B\kern-.05em{\sc i\kern-.025em b}\kern-.08em
    T\kern-.1667em\lower.7ex\hbox{E}\kern-.125emX}}

\title{Peripheral prototypical contrastive loss\\ for supervised medical image segmentation}
\name{Takuro Shimaya \qquad Ryosaku Shino \qquad Masahiro Saiko}
\address{Biometrics Research Laboratories, NEC Corporation, Kawasaki, Japan}

\begin{document}

\maketitle

\begingroup
\renewcommand{\thefootnote}{}
\footnotetext{%
\textcopyright~2026 IEEE. Personal use of this material is permitted.
Permission from IEEE must be obtained for all other uses,
in any current or future media, including reprinting/republishing
this material for advertising or promotional purposes,
creating new collective works, for resale or redistribution
to servers or lists, or reuse of any copyrighted component
of this work in other works.
}
\endgroup

\begin{abstract} 
Robust segmentation of biomedical images remains challenging primarily due to ill-defined object boundaries.
Although several loss functions have recently been proposed to enhance object boundary recognition, concerns remain about the susceptibility to unstable annotations resulting from boundary ambiguities.
Here, we introduce the Peripheral Prototypical Contrastive (PPC) loss, which leverages prototypical contrastive learning to improve boundary recognition and enable stable training even under noisy annotation conditions.
The PPC loss utilizes prototype features obtained from local regions near object boundaries, focusing specifically on ambiguous areas that have been challenging for existing approaches.
Extensive experiments across multiple datasets demonstrated that the PPC loss consistently improve segmentation performance even under conditions with noisy labels.
A newly proposed strategy, which adaptively determines the local region for prototype calculation, enhanced robustness and outperformed existing methods.
Our findings indicate the potential of the PPC loss in facilitating scalable and reliable medical image segmentation, including the development of foundation models with imperfect annotations.
\end{abstract}
\section{Introduction}
\label{sec:Intro}
Segmentation has long been a crucial task in biomedical image analysis, and its methodologies have continued to advance.
In recent years, a continuous stream of foundation models targeting medical imaging has been proposed\cite{Cheng2023,Ma2024,Zhao2025}, demonstrating that the utilization of large-scale datasets can address many challenges that are specific to the medical domain.
Nevertheless, several critical issues still persist despite the rise of these foundation models.
For instance, lesions and organs frequently exhibit ambiguous and ill-defined margins, which can introduce several complications such as annotation noise \cite{Kumari2025} and degradation in object boundary recognition of the foundation model \cite{Cheng2023}.
To address these challenges, various approaches have been explored, including specialized neural network architecture designs \cite{Lee2020,Wang2023}, the utilization of labels from multiple annotators \cite{Kumari2025,Rahman2023}, and proposals for novel loss functions \cite{Huang2024,Zhang2024}.
Among these, the development of novel loss functions for enhancing boundary recognition can be highly versatile and, as noted in ref.\cite{Cheng2023}, holds promise for further improving foundation models.

Several loss functions have been investigated to improve boundary identification.
In the context of generic images, camouflaged object detection (COD) \cite{Xiao2024, He2023} addresses similar challenges, aiming to segment objects whose textures are similar to those of the background.
For the COD task, the multi-task learning framework has been commonly applied, in which the model is simultaneously trained to segment boundary pixels using the edge detection loss \cite{Xiao2024, He2023}. 
In the biomedical field, contour-weighted cross-entropy and Dice (CWCD) loss \cite{Huang2024} and ROI-based contrastive loss \cite{Zhang2024} have been proposed.
CWCD loss amplifies cross-entropy and Dice losses near object contours to enhance boundary recognition.
ROI-based contrastive loss applies instance-wise contrastive (IC) learning to pixel-wise features randomly sampled around object boundaries, promoting the separation of features around boundaries.

Although the above-mentioned existing methods can be effective when annotations are well-defined and clean, the loss functions can be highly perturbed when annotations are unstable or inconsistent, possibly degrading segmentation performance.
The values of the edge detection loss and CWCD loss vary significantly depending on the position of object boundaries, resulting in high susceptibility to annotation errors.
Conversely, contrastive learning induces a relative separation of pixel-wise features between foreground and background, without enforcing abrupt changes in the features around object boundaries.
While this seems to improve robustness against label noise, it has been pointed out that IC learning underperform when incorrectly assigned negative pairs are present \cite{Li2021}, which can be caused by pixel-level annotation errors in medical images.

In this study, we design a prototype-based loss to further enhance robustness against ambiguous object boundaries.
Prototypical contrastive (PC) learning \cite{Li2021} incorporates representative features of each class, i.e., prototypes, and has already been applied to segmentation tasks \cite{Du2022,He2025}.
While IC learning utilizes pixel-wise features to obtain the contrastive loss, PC learning leverages the mean of pixel-wise features belonging to the same class.
Unlike pixel-wise features, the prototypical feature obtained thereby is less affected by pixel-level perturbation of annotation masks, leading to more stable training than IC learning.
Although PC learning has been particularly utilized in weakly-supervised settings \cite{Wu2024,Wang2024}, few studies have explored the enhancement of object boundary recognition in fully-supervised settings.
Here we develop the peripheral prototypical contrastive (PPC) loss, where the prototype is computed using pixel-wise features from hard negative pixels around the object boundary.
Our work yields the following contributions:
\begin{enumerate}
\item We introduce the idea of PC learning instead of IC learning to enhance boundary recognition in fully-supervised medical images with both clean and noisy annotations.
\item We propose a method for adaptively determining the peripheral region for prototype calculation, which further improves the segmentation performance.
\item Through a series of experiments on three datasets, our PPC loss demonstrates the potential applicability across various imaging modalities and robustness against label noise.
\end{enumerate}
\begin{figure}[t]
    \centerline{\includegraphics[width=\linewidth]{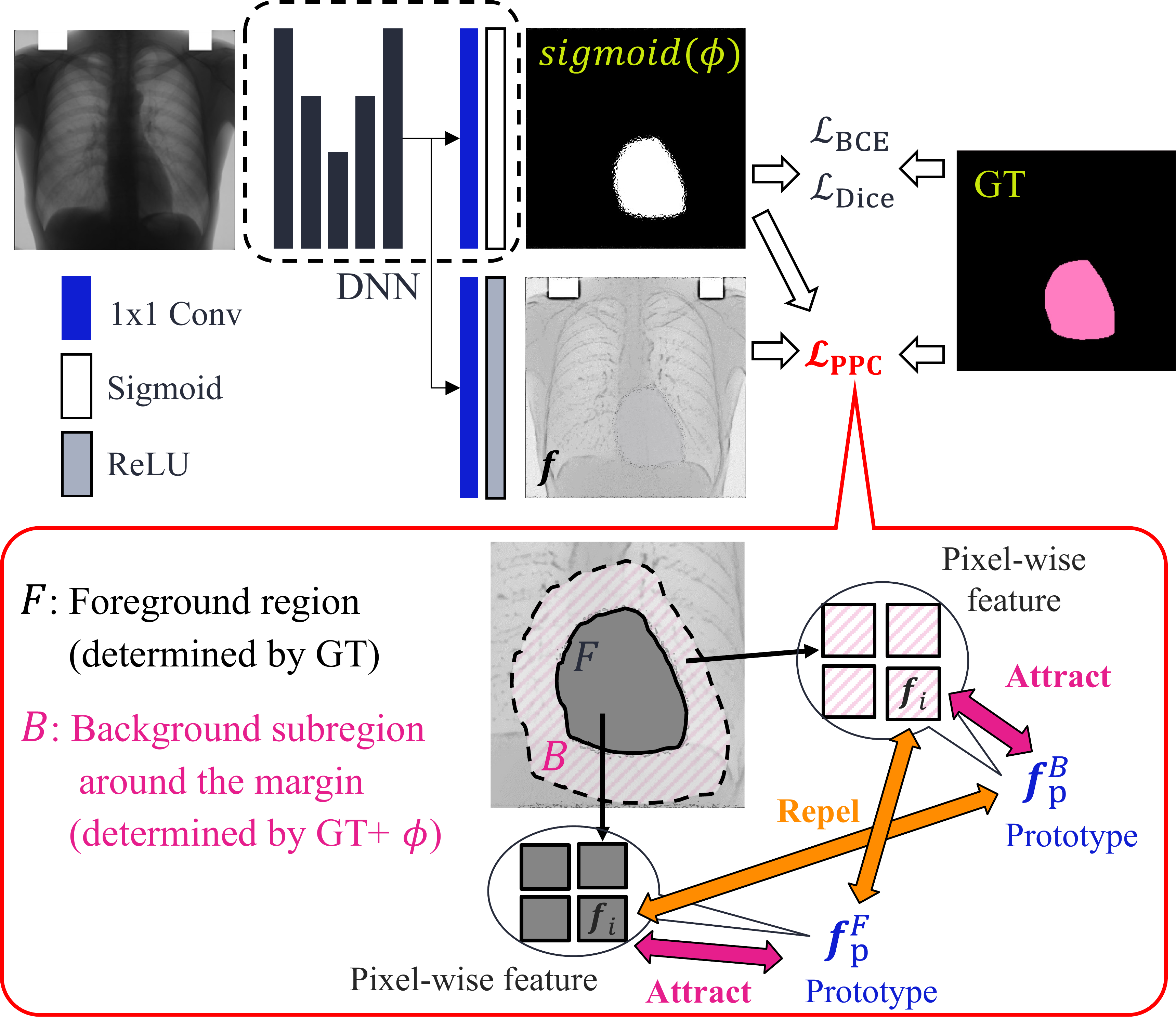}}
    \caption{The overview of the proposed the PPC loss ($\mathcal{L}_\mathrm{PPC}$) to enhance boundary recognition of the DNN.
    }
    \label{fig:1}
\end{figure}

\section{Methods}
\label{sec:Method}
\subsection{PPC loss}
The key concept of our proposed PPC loss lies in performing prototypical contrastive learning using pixel-wise features extracted from the local regions near object boundaries.
Among background regions, which often occupy a large portion of medical images, subregions close to the target foreground $F$ are particularly hard to discriminate, and are pivotal for training segmentation models.
We therefore define a background subregion $B$ around the margin of $F$, and propose the PPC loss for an image $n$:
\begin{align}
    \label{eq:1}
    \mathcal{L}^n_\mathrm{PPC}&=-\left[ \frac{1}{|B|}\sum_{i\in B}\log\lrparen{\frac{e^{S(\bm{f}_i,\bm{f}^B_\mathrm{p})/\tau}}{e^{S(\bm{f}_i,\bm{f}^B_\mathrm{p})/\tau}+e^{S(\bm{f}_i,\bm{f}^F_\mathrm{p})/\tau}}} \right. \nonumber \\
    & \left. +\frac{1}{|F|}\sum_{i\in F}\log\lrparen{\frac{e^{S(\bm{f}_i,\bm{f}^F_\mathrm{p})/\tau}}{e^{S(\bm{f}_i,\bm{f}^F_\mathrm{p})/\tau}+e^{S(\bm{f}_i,\bm{f}^B_\mathrm{p})/\tau}}} \right],
\end{align}
where $\bm{f}_i$, $\tau$, $S(\cdot,\cdot)$, and $|\cdot|$ denote the feature at pixel $i$, the temperature constant, the cosine similarity, and the number of pixels in the specified region, respectively (see \figref{fig:1}).
We fixed $\tau = 1$ following ref.\cite{Du2022}.
The pixel-wise features for contrastive learning are obtained by applying a 1×1 convolutional layer, followed by a ReLU activation, to the features output by the deep neural network (DNN) under training.
$\bm{f}^{\{B,F\}}_\mathrm{p}$ is the prototype, defined as the mean feature vector within the respective region.
We train the DNN using $\mathcal{L}^n = \mathcal{L}^n_\mathrm{BCE} + \lambda_\mathrm{D}\mathcal{L}^n_\mathrm{Dice} + \lambda_\mathrm{P}\mathcal{L}^n_\mathrm{PPC}$, where $\mathcal{L}^n_\mathrm{BCE}$, $\mathcal{L}^n_\mathrm{Dice}$, and $\lambda_\mathrm{\{D,P\}}$ are binary cross-entropy (BCE) loss, Dice loss, and the weighting constants, respectively. 

The simplest definition of the background subregion $B$ is the set of all background pixels in the image: $\hat{B}_\mathrm{All} = \{i\mid i \notin F\}$, which yields the existing PC loss with the background class as defined by the given annotation.
To obtain boundary-aware prototypes, the most straightforward way is to define a local region isotropically surrounding $F$, which has been implemented in refs.\cite{Wu2024,Wang2024} for weakly-supervised situations.
Specifically, letting $r_i$ denote the distance from pixel $i$ to the boundary of $F$, one can define $B$ as
\begin{align}
    \label{eq:2}
    \hat{B}_\mathrm{Iso} = \{i\mid r_i < \delta \cap i \notin F\}
\end{align}
with a hyperparameter for the threshold distance $\delta$.
Nevertheless, as the subregion $B$ is not necessarily isotropic in all cases, a more adaptive approach is desired.
To address this, we propose defining $B$ using the logit outputs $\phi_i$ from the DNN during training (\figref{fig:1}).
We expect that the background pixels whose $\phi_i$ is close to 0, satisfying $\phi_i+|\lrbracket{\phi_{i'}}_{r_{i'}\approx \delta}|>0$, are difficult to discriminate and are distributed around $F$ with an approximate width of $\delta$.
Here $\lrbracket{\phi_{i'}}_{r_{i'}\approx \delta}$ denotes the mean logit over the background pixels where $r_i \approx \delta$.
However, such a simple additive operation is so susceptible to noise in $\phi$ that the region specified above might include pixels distant from $F$.
We mitigate this by introducing a variance parameter $\sigma$ and ultimately obtain
\begin{align}
    \label{eq:3}
    &\hat{B}_\mathrm{Ada} = \nonumber \\
    &\{i\mid \phi_i+|\lrbracket{\phi_{i'}}_{r_{i'}\approx \delta}|e^{-(r_i/2\sigma)^2+(\delta/2\sigma)^2}>0 \cap i \notin F\}.
\end{align}
The coefficient $e^{-(r_i/2\sigma)^2+(\delta/2\sigma)^2}$ of the additive term follows a Gaussian profile, taking larger values for pixels closer to $F$ and becomes 1 when $r_i = \delta$.
This allows for the adaptive determination of $B$ on a per-case basis.
\begin{table*}[t]
\centering
\small
\begin{tabular}{l|cc|cc|cc|cc|cc}
 & \multicolumn{6}{c|}{Chest X-ray (JSRT)} & \multicolumn{2}{c|}{CT (LIDC-IDRI)} & \multicolumn{2}{c}{MR (BraTS2020)} \\
 & \multicolumn{2}{c|}{Lung} & \multicolumn{2}{c|}{Heart} & \multicolumn{2}{c|}{Clavicle} & \multicolumn{2}{c|}{Lung nodule} & \multicolumn{2}{c}{Brain tumor} \\
Methods & Clean & Noise & Clean & Noise & Clean & Noise & Clean & Noise & Clean & Noise \\
\midrule
BCE loss + Dice loss   & 97.50 & 93.45 & 89.77 & 80.88 & 86.25 & 77.91 & 64.21 & 59.39 & 83.35 & 81.18 \\
(Baseline)             & $\pm$0.05 & $\pm$0.25 & $\pm$1.40 & $\pm$1.70 & $\pm$1.15 & $\pm$1.09 & $\pm$0.40 & $\pm$0.72 & $\pm$0.29 & $\pm$0.21 \\
\hline
\multirow{2}{*}{+ Edge detection loss} & 97.75 & \cellcolor{lightgray}93.44 & \cellcolor{lightgray}86.02 & \cellcolor{lightgray}75.09 & \cellcolor{lightgray}86.06 & 78.92 & 65.06 & 59.50 & \cellcolor{lightgray}82.77 & \cellcolor{lightgray}78.99 \\
                     & $\pm$0.03 & \cellcolor{lightgray}$\pm$0.14 & \cellcolor{lightgray}$\pm$3.74 & \cellcolor{lightgray}$\pm$3.41 & \cellcolor{lightgray}$\pm$0.41 & $\pm$0.89 & $\pm$0.33 & $\pm$0.58 & \cellcolor{lightgray}$\pm$0.66 & \cellcolor{lightgray}$\pm$0.39 \\
\multirow{2}{*}{+ CWCD loss} & 97.50 & \cellcolor{lightgray}91.08 & 92.02 & 81.36 & 87.29 & \cellcolor{lightgray}74.18 & \cellcolor{lightgray}60.94 & \cellcolor{lightgray}54.53 & \cellcolor{lightgray}83.27 & \cellcolor{lightgray}80.67 \\
                           & $\pm$0.09 & \cellcolor{lightgray}$\pm$0.23 & $\pm$0.50 & $\pm$1.46 & $\pm$0.72 & \cellcolor{lightgray}$\pm$1.35 & \cellcolor{lightgray}$\pm$0.35 & \cellcolor{lightgray}$\pm$0.37 & \cellcolor{lightgray}$\pm$0.38 & \cellcolor{lightgray}$\pm$0.38 \\
+ ROI-based                 & \cellcolor{lightgray}97.49 & \cellcolor{lightgray}91.81 & 93.87 & 87.62 & 90.88 & 79.67 & 64.35 & \cellcolor{lightgray}57.90 & \uline{83.82} & \textbf{82.06} \\
\hspace{5pt} contrastive loss          & \cellcolor{lightgray}$\pm$0.03 & \cellcolor{lightgray}$\pm$0.49 & $\pm$0.07 & $\pm$0.31 & $\pm$0.16 & $\pm$2.05 & $\pm$1.91 & \cellcolor{lightgray}$\pm$1.19 & \uline{$\pm$0.23} & $\pm$\textbf{0.14} \\
\multirow{2}{*}{+ PC loss}  & \uline{97.84} & \cellcolor{lightgray}93.30 & 94.00 & 88.80 & 90.32 & 79.71 & 64.52 & \cellcolor{lightgray}58.88 & 83.62 & \cellcolor{lightgray}80.53 \\
                          & \uline{$\pm$0.03} & \cellcolor{lightgray}$\pm$0.30 & $\pm$0.17 & $\pm$0.48 & $\pm$0.25 & $\pm$0.91 & $\pm$0.42 & \cellcolor{lightgray}$\pm$0.88 & $\pm$0.15 & \cellcolor{lightgray}$\pm$0.34 \\
                          + Isotropic PPC loss  & \uline{97.84} & \textbf{94.44} & \uline{94.09} & \uline{89.33} & \uline{91.66} & \uline{86.81} & \uline{65.30} & \uline{60.64} & \cellcolor{lightgray}83.24 & 81.62 \\
\hspace{5pt} ($B=\hat{B}_\mathrm{Iso}$) & \uline{$\pm$0.03} & $\pm$\textbf{0.25} & \uline{$\pm$0.14} & \uline{$\pm$0.29} & \uline{$\pm$0.17} & \uline{$\pm$0.62} & \uline{$\pm$0.25} & \uline{$\pm$0.62} & \cellcolor{lightgray}$\pm$0.34 & $\pm$0.24 \\
+ \textbf{Adaptive PPC loss (ours)}  & \textbf{97.96} & \uline{93.96} & \textbf{94.54} & \textbf{89.85} & \textbf{92.27} & \textbf{87.32} & \textbf{66.04} & \textbf{61.84} & \textbf{84.52} & \uline{81.92} \\
\hspace{5pt} ($\bm{B=\hat{B}_\mathrm{Ada}}$) & $\pm$\textbf{0.01} & \uline{$\pm$0.32} & $\pm$\textbf{0.05} & $\pm$\textbf{0.26} & $\pm$\textbf{0.18} & $\pm$\textbf{0.70} & $\pm$\textbf{0.57} & $\pm$\textbf{0.47} & $\pm$\textbf{0.18} & \uline{$\pm$0.23} \\\bottomrule
\end{tabular}
\caption{Comparison of segmentation performance using the Dice coefficient. The best and second-best scores are highlighted in \textbf{bold} and \uline{underlined}, respectively. Gray cells indicate the degradation compared to the baseline.}
\label{tab:1}
\end{table*}

\subsection{Experiments}
We adopted U-Net \cite{Ronneberger2015} and 3D U-Net \cite{Cicek2016} as DNNs in \figref{fig:1} and compared the performance of the PPC loss to existing methods.
We used the baseline using only BCE loss and Dice loss, and evaluated each method by additionally incorporating edge detection loss \cite{He2023}, CWCD loss \cite{Huang2024}, ROI-based contrastive loss \cite{Zhang2024}, PC loss ($\mathcal{L}_\mathrm{PPC}|_{B=\hat{B}_\mathrm{All}}$) \cite{Du2022}, the isotropic PPC loss ($\mathcal{L}_\mathrm{PPC}|_{B=\hat{B}_\mathrm{Iso}}$) \cite{Wang2024}, and the adaptive PPC loss ($\mathcal{L}_\mathrm{PPC}|_{B=\hat{B}_\mathrm{Ada}}$).
In addition, to verify the stability of our method against annotation noise, we performed experiments not only with clean labels, but also with noisy labels during the training and validation phases.
During the test phase, clean labels were always used as ground truth (GT).
We performed five independent training runs and evaluated the performance using the mean and standard error of the Dice coefficient.
The hyperparameters were independently optimized for each experiment using grid search, within $\lambda\in[0.1,1]$, $\delta\in[3, 30]$, and $\sigma\in[2, 20]$.

Experiments were conducted using three publicly available datasets as follows.
The first is the JSRT database \cite{Shiraishi2000}, consisting of 247 chest X-ray images with corresponding masks for the lungs, heart, and clavicles.
Data preprocessing and splitting were performed according to ref.\cite{Yao2023}.
Noisy labels were artificially generated by randomly dilating or eroding the masks following the methods in refs.\cite{Zhang2020,Yao2023}.
We trained and evaluated models independently for each class.
The second dataset is LIDC-IDRI \cite{Armato2011}, which we extracted and processed to obtain 13,085 two-dimensional slices of CT with lung nodule masks according to refs.\cite{Yao2023,Shimaya2024}.
This dataset includes annotations from multiple experts, allowing for the use of natural label noise.
Specifically, clean labels were obtained by majority voting across all annotation masks, while a randomly selected single mask was used as the noisy label.
The third dataset is BraTS2020 \cite{Menze2015,Bakas2017,Bakas2018}, which includes 369 three-dimensional MRI images of the brain and corresponding tumor masks.
Data preprocessing and splitting were conducted following ref.\cite{Yao2023}, and label noise was generated in the same manner as in the JSRT database.
U-Net was adopted for two-dimensional segmentation tasks on the JSRT database and LIDC-IDRI, while 3D U-Net was used for BraTS2020.

\section{Results}
\label{sec:Results}
\begin{figure}[t]
    \centerline{\includegraphics[width=\linewidth]{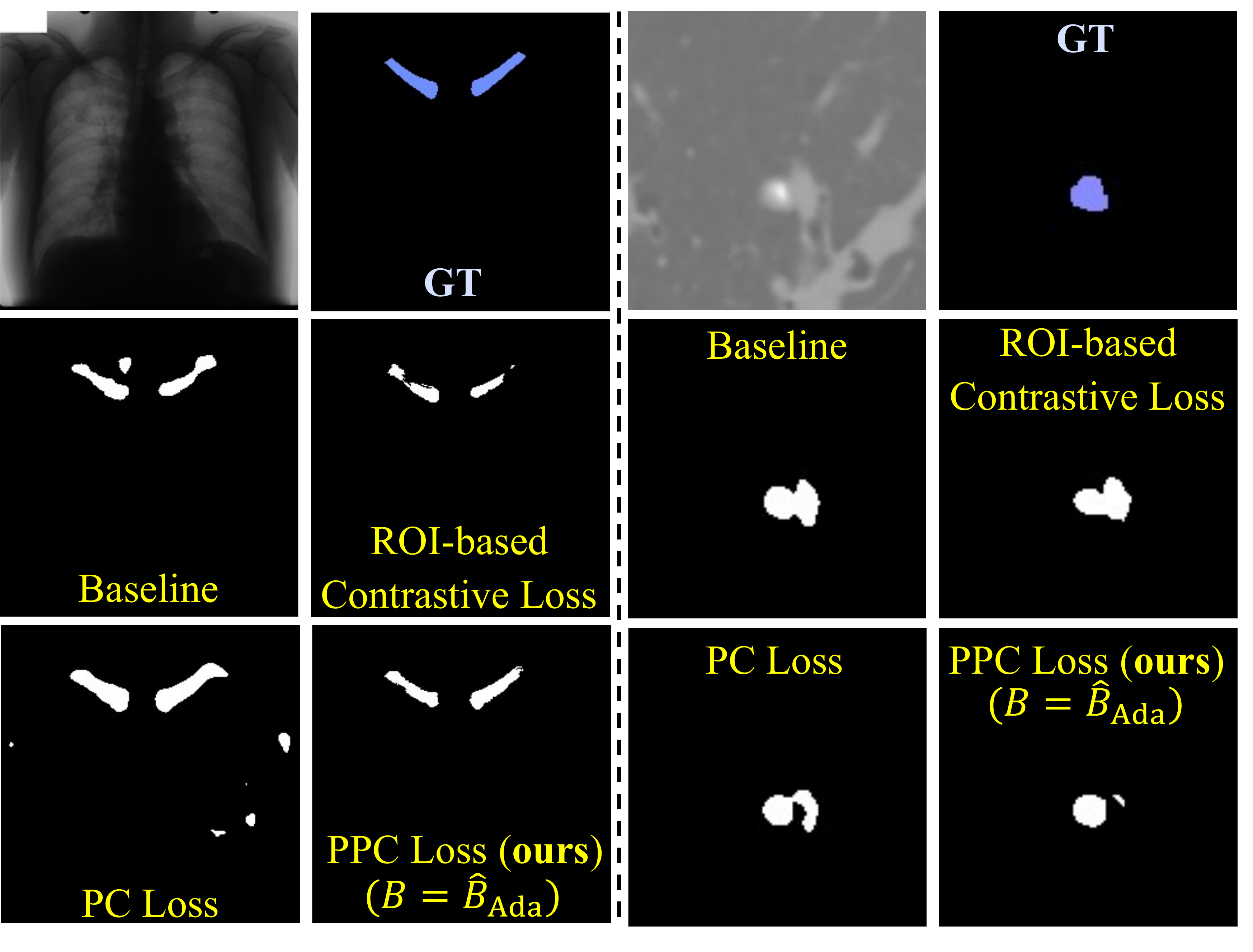}}
    \caption{Visual comparison of predictions by the DNN trained with noisy labels.
    The ground truth (GT) is shown in blue.
    (Left) Results for the clavicles in the JSRT database.
    (Right) Results for the nodule in LIDC-IDRI.
    }
    \label{fig:2}
\end{figure}

As shown in \tabref{tab:1}, the DNN trained with the adaptive PPC loss demonstrated the best performance in the majority of cases.
Notably, significant improvement was achieved for clavicles in chest X-rays and lung nodules in CT.
A visual comparison of the segmentation results for the aforementioned cases (\figref{fig:2}) shows particularly remarkable differences at ambiguous boundaries, such as at both ends of clavicles and at the interface between normal tissue and nodules.

Regarding previous methods, the Dice coefficient frequently dropped below that of the baseline when using noisy labels for training (see gray cells in \tabref{tab:1}).
This suggests that existing methods, particularly wituout prototypical contrastive learning, are not sufficiently robust to annotation noise.
In contrast, only our adaptive PPC loss exhibited the stable performance without a decrease in the Dice coefficient even when using noisy labels.

\begin{figure}[t]
    \centerline{\includegraphics[width=\linewidth]{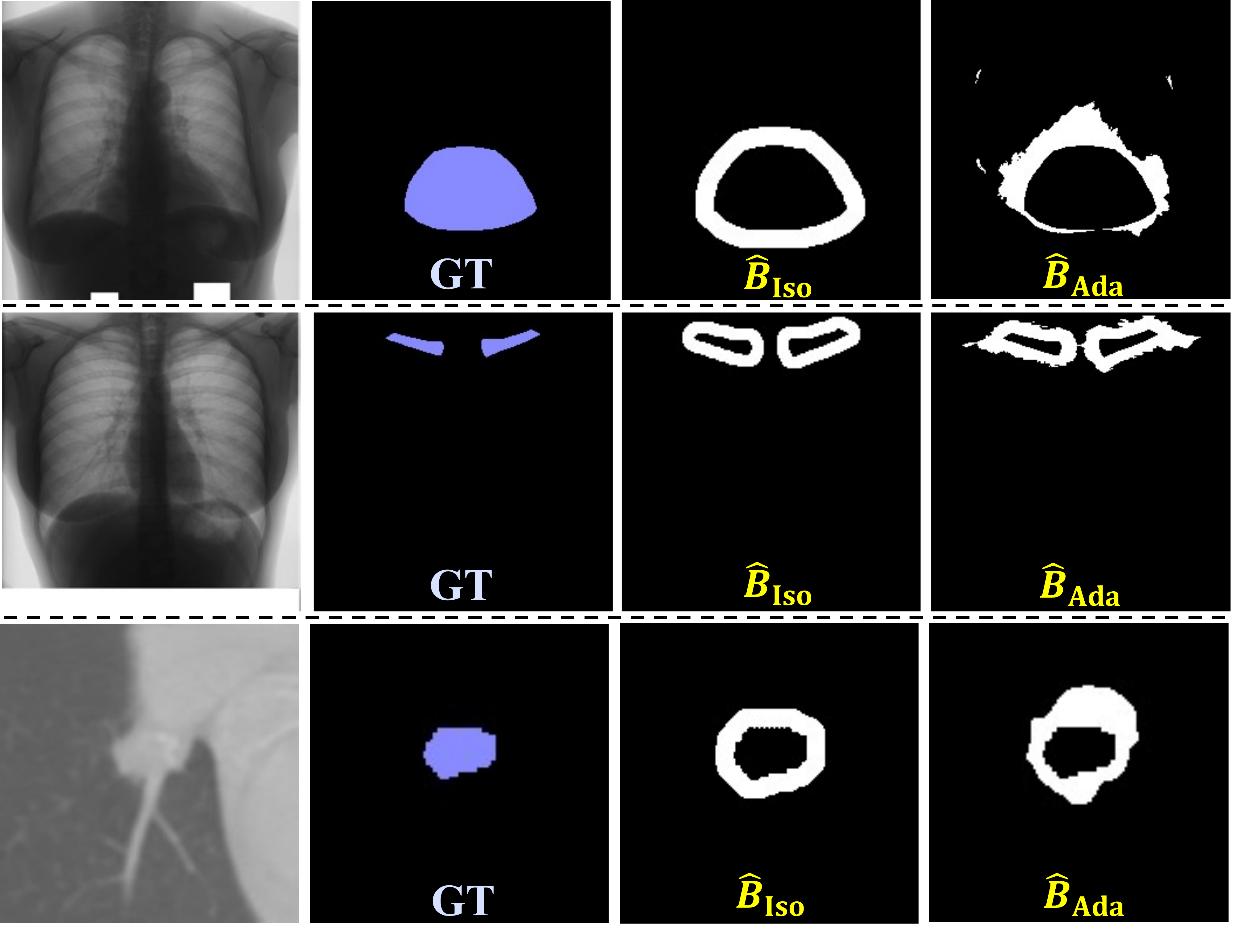}}
    \caption{Visual comparison of the background subregions $B$.
    $\hat{B}_\mathrm{Ada}$ was estimated by the DNN trained with noisy labels.
    (Top) Results for the heart in the JSRT database.
    (Middle) Results for the clavicles in the JSRT database.
    (Bottom) Results for the nodule in LIDC-IDRI.
    }
    \label{fig:3}
\end{figure}

The definition of the background subregion $B$ indeed significantly affected segmentation performance (\tabref{tab:1}).
We visually compared $\hat{B}_\mathrm{Ada}$, which was derived using the logit from the DNN (\equaref{eq:3}) after training had converged, with $\hat{B}_\mathrm{Iso}$, which is defined isotropically with a fixed $\delta$ (\equaref{eq:2}).
$\hat{B}_\mathrm{Ada}$ exhibited anisotropic spreading, e.g., along the periphery of the heart, around regions where the clavicle overlaps with the pleura, and at the boundaries between nodules and normal tissue (\figref{fig:3}).
This suggests that $\hat{B}_\mathrm{Ada}$ adaptively and selectively incorporates ambiguous regions into the calculation of the PPC loss.

\section{Discussion and Conclusion}
\label{sec:Disc}
In summary, we developed the PPC loss to enable robust training of segmentation models even in the presence of ambiguous object boundaries and unstable annotations in medical images.
Across multiple datasets and tasks exhibiting diverse characteristics, our proposed method consistently outperformed the baseline in every experiment.

The novelty of the PPC loss lies in its application of prototypical contrastive learning to improve the capability for recognizing ambiguous object boundaries in fully-supervised settings with both clean and noisy annotations.
In instance-wise contrastive learning, loss functions are computed by referencing randomly selected pixel-wise features as positive and negative pairs for the anchor.
In the presence of label noise or ill-defined boundaries, randomly sampled pixel-wise features are likely to form inappropriate positive or negative pairs.
In contrast, the PPC loss utilizes the prototype, i.e., the mean feature within each class, as the reference for the anchor, making it robust to pixel-level perturbations in annotations.
Furthermore, by adaptively detecting boundary-aware ambiguous regions for prototype estimation, the PPC loss successfully achieved stable training of the DNN.

Our method offers several avenues for further performance improvement.
First, the PPC loss cannot handle all types of annotation noise.
For example, the PPC loss may not be effective when annotations are consistently larger or smaller than the true labels.
In such situations, our method should be combined with, e.g., label correction methods \cite{Zhang2020,Yao2023,Shimaya2024}.
Second, the adaptive estimation of background subregion $B$ requires two hyperparameters, $\delta$ and $\sigma$.
The optimal hyperparameter values might vary across different target types, potentially increasing the burden when training multi-class segmentation models.
It is therefore necessary to explore solutions, such as making these hyperparameters learnable or establishing a theoretical framework to determine them analytically.
Third, while this study conducted prototypical contrastive learning within a single image, we might be able to compute the loss over multiple images, as demonstrated in ref.\cite{Zhang2024}.
Lastly, this study used all pixels in the foreground $F$ to obtain the PPC loss since most target objects were sufficiently small.
Segmentation performance may be further improved by extracting the foreground subregion when $F$ is sufficiently large, in a manner similar to $B$. 

Beyond these methodological considerations, we also highlight the broader potential applicability of our method.
For instance, while this study evaluated performance on binary segmentation for a specific target, the PPC loss may benefit the training of foundation models.
One can simply incorporate our method by adding the PPC loss to the existing BCE or Dice loss, regardless of the network architecture, as long as feature maps can be obtained.
Moreover, recent foundation models often utilize pseudo-labels during training \cite{Cheng2023,Ma2024}, making it increasingly challenging to maintain the quality of teacher labels.
The PPC loss is, therefore, expected to be even more effective when training foundation models with large numbers of noisy labels.

\subsection*{Compliance with ethical standards}
Ethical approval was not required, as confirmed by the license attached with the open data that we used.

\subsection*{Acknowledgments}
The authors are employees of NEC corporation.

\ninept
\bibliographystyle{IEEEbib}
\bibliography{paper}

\end{document}